\documentclass[runningheads]{llncs}
\usepackage[T1]{fontenc}
\usepackage{graphicx,verbatim}
\usepackage{amsfonts}
\usepackage{amsmath}
\usepackage{amssymb}
\usepackage{booktabs}
\usepackage{hyperref}
\usepackage{multirow}
\usepackage{pifont}
\newcommand{\cmark}{\ding{51}}%
\newcommand{\xmark}{\ding{55}}%

\usepackage{xcolor}
\newcommand\methodname{DualTrack}

\begin{document}
\title{\methodname{}: Sensorless 3D Ultrasound needs Local and Global Context}
%
\author{Paul F. R. Wilson\inst{1} \and
Matteo Ronchetti\inst{2} \and
R\"udiger G\"obl\inst{2} \and
Viktoria Markova\inst{2} \and 
Sebastian Rosenzweig\inst{2} \and 
Raphael Prevost\inst{2} \and
Parvin Mousavi\inst{1} \and 
Oliver Zettinig\inst{2}
\authorrunning{P.F.R. Wilson et al.}
%
\institute{Queen's University, Kingston, Canada \and
ImFusion GmbH, Munich, Germany \\
\email{paul.wilson@queensu.ca}\\}
}
    
\maketitle              
\begin{abstract}

Three-dimensional ultrasound (US) offers many clinical advantages over conventional 2D imaging, yet its widespread adoption is limited by the cost and complexity of traditional 3D systems.
Sensorless 3D US, using deep learning to estimate a 3D probe trajectory from a sequence of 2D US images, is a promising alternative. 
\emph{Local features} such as speckle patterns can help predict frame-to-frame motions, while \emph{global features}, such as coarse shapes and anatomical structures, can situate the scan relative to anatomy and help predict its general shape. In prior approaches, global features are either ignored or tightly coupled with local feature extraction, restricting the ability to robustly model these two complementary aspects. We propose \methodname{}, a novel dual encoder architecture leveraging decoupled \emph{local} and \emph{global} encoders specializing in their respective scale of feature extraction. The local encoder uses dense spatiotemporal convolutions to capture fine-grained features, while the global encoder utilizes an image backbone such as a 2D CNN or foundation model and temporal attention layers to embed high-level anatomical features and long-range dependencies. A lightweight fusion module then combines these features to estimate trajectory. Experimental results on a large public benchmark show that \methodname{} achieves state-of-the-art accuracy and globally consistent 3D reconstructions, outperforming previous methods and yielding an average reconstruction error below 5~$mm$.\footnote{Code and model weights are provided at \url{https://github.com/ImFusionGmbH/DualTrack}.}


\end{abstract}

\section{Introduction}

3D ultrasound (US) provides significant advantages over conventional 2D imaging, including enhanced visualization, improved anatomical navigation, and accurate volumetric measurements. 3D US systems that use matrix probes or external tracking for volume construction can be cost prohibitive and complex, or impose physical constraints that are impractical in routine clinical workflows. Sensorless 3D US, approaches that use deep learning to directly estimate the probe's 3D trajectory from 2D image sequences, are a promising alternative. 

\begin{figure}[t]
    \centering
    \includegraphics[width=1.0\linewidth]{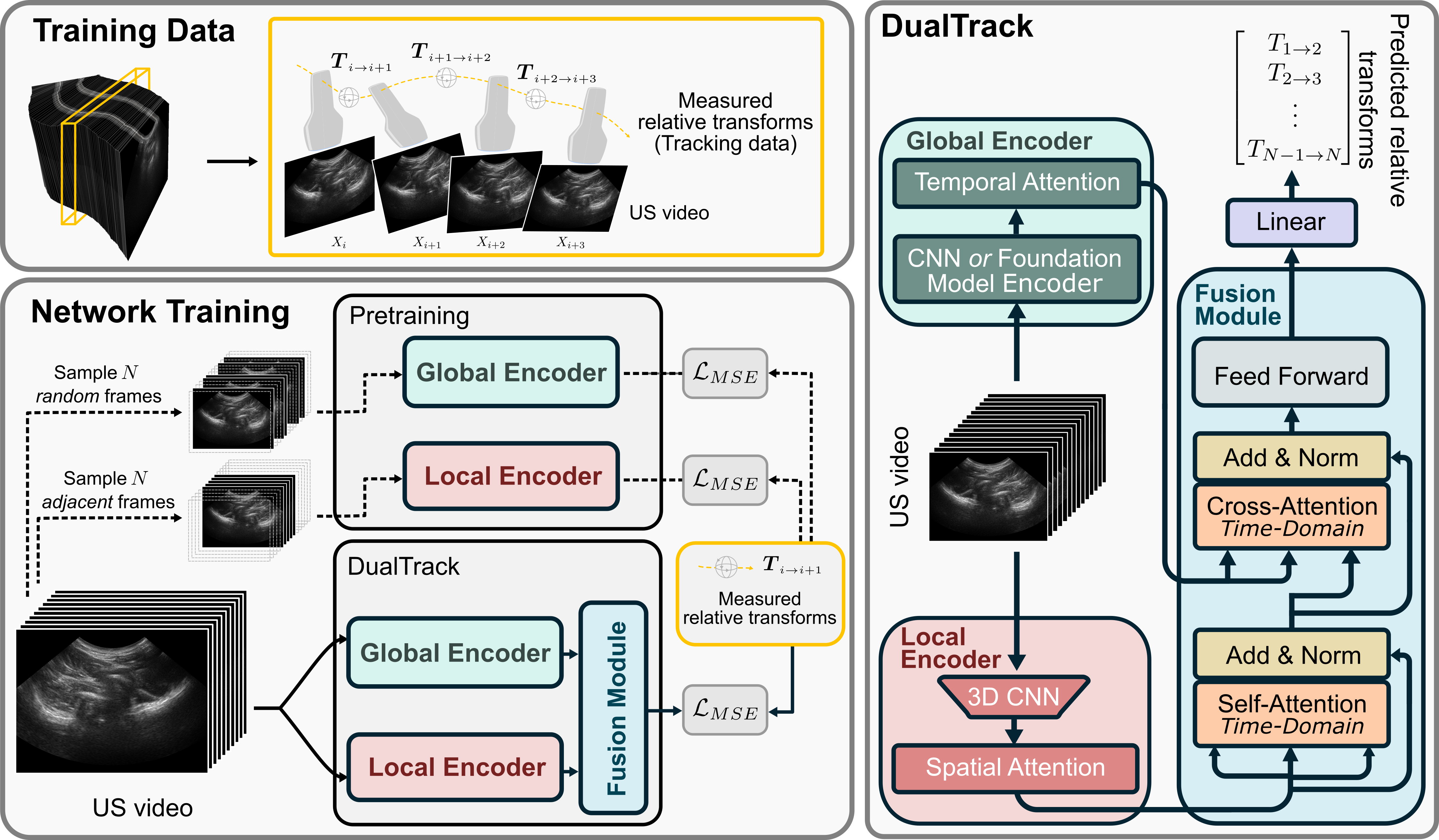}
    \caption{\methodname{} enables sensorless 3D ultrasound by estimating the probe trajectory for a sequence of ultrasound images (Top Left). Specialized local and global encoders are designed to extract low- and high-level features--their outputs are combined using self- and cross-attention through time (Right). The encoders are pretrained at their respective timescales before being finetuned with the fusion module (Bottom Left).}
    \label{fig:method}
\end{figure}

A sequence of 2D US images contains rich spatiotemporal information that can, in principle, be leveraged to infer the probe’s 3D position. \emph{Local features}, such as speckle patterns and short-term frame-to-frame motion cues, capture fine-grained displacements between adjacent frames. Meanwhile, \emph{global features}, derived from identifying coarse anatomical patterns in the image sequence, help situate the probe with respect to that anatomy. For example, identifying structures such as bones, vessels, muscles, and organs in a 2D image sequence, and deriving their 3D properties such as continuity, shape, and landmarks can help inform a realistic 3D reconstruction that is faithful to the original structure. Therefore, an ideal trajectory estimation model should combine local \emph{and} global features to ensure both short-term accuracy and global consistency. In practice, robustly modeling these complementary aspects remains a significant challenge.

Early advances in sensorless 3D US using convolutional neural networks (CNNs) were driven primarily by improvements in \emph{local} feature extraction. These approaches successfully capture low-level image details such as speckle patterns and motion cues~\cite{guo2020sensorless,guo2023,li2023trackerless,prevost20183d}, but are inherently blind to global features. Recognizing this shortcoming, hybrid architectures have been proposed that append a sequence model (e.g. long short-term memory (LSTM) or transformer) to a CNN, enabling global feature extraction by capturing long-term temporal dependencies~\cite{luo2021,luo2022deep,ning2022spatial}. Loss functions aimed at enforcing global consistency have also been used~\cite{li2023trackerless,luo2022deep}. The limitation of these approaches is their strong coupling: the same basic encoder is forced to model local and global features, even though the properties of an ideal local feature extractor (locality, texture bias) do not match those of a global feature extractor (shape, long-range dependency). 

In this work, we propose a novel approach for sensorless 3D US that maximizes the learning of complementary local and global information from 2D US sequences. Our approach is inspired by the \emph{dual encoder} paradigm, i.e. a network architecture comprised of two separate branches, designed to model different features of an image--e.g., spatial vs. frequency features~\cite{farshad2022net} or texture vs. shape features~\cite{fu2022deau,zhang2021transfuse,hong2023dual}. This separation allows for complementary and specialized encoders for each type of feature. We extend the dual encoder paradigm to 3D US by designing specialized spatiotemporal encoders that capture local and global properties. The decoupled nature of our architecture allows us to explore specialized design choices in each encoder, such as the use of CNNs for local feature modeling and pretraining or transfer learning (e.g. self-supervised learning, foundation models) to improve global feature modeling. After an initial independent training phase, these encoders are combined using a fusion module that integrates the outputs of each encoder to predict trajectory.

Our approach,  \methodname, is the first (to our knowledge) dual encoder network for sensorless 3D US. Our main contributions are:
\begin{enumerate} 
\item We propose a dual encoder architecture that separately models local features (e.g. speckle-based cues) and global features (e.g. high-level anatomical landmarks), before combining them using a fusion module and prediction head to generate accurate and globally consistent trajectory estimates. 
\item We design specialized architectures and pretraining objectives for each encoder to maximize its performance in its designated role. In particular, we investigate the potential of medical foundation models in sensorless 3D US by incorporating them as part of the global encoder.

\item Using a large public benchmark dataset, we empirically validate our method. \methodname{} significantly outperforms previous approaches, achieving an average global reconstruction error of less than 5 $mm$. 
\end{enumerate}

\section{Methods}

An overview of our method is shown in Figure~\ref{fig:method}. We train a network to estimate probe trajectory from a sequence of 2D US images. Our network consists of a local and a global encoder, designed to extract low- and high-level features, respectively. The networks are pre-trained at their respective time scales. Finally, we train a fusion module to combine embeddings from the local and global encoders and produce accurate, globally consistent trajectory estimates.

\noindent\textbf{Problem setup:} Given an ordered sequence of 2D US frames $(X_i)_{i=1}^{n}$, the objective is to reconstruct the trajectory of the US probe by estimating its pose at each timepoint $i$. The pose at timepoint $i$ can be represented by a homogeneous $4\times4$ matrix $T_i$ or the corresponding six degree of freedom (three translations and three Euler angles) parameterization $\mathbf{p}_i \in \mathbb{R}^6$.
Following prior approaches~\cite{prevost20183d,li2023trackerless}, we set $T_0 = \text{Id}$ without loss of generality and focus on estimating the relative transformations $T_{j \leftarrow i} = T_{i}^{-1} T_j$, represented by 6-dimensional parameter vectors $\mathbf{p}_{j \leftarrow i}$. This reduces to a deep regression problem which we solve using a parameterized network $f(\cdot\:; \Theta)$ such that $f\big((X_i)_{i=1}^{n}  ; \Theta \big) \approx (\mathbf{p}_{j \leftarrow i})_{i=1}^{n-1}$ by optimizing the mean-squared error loss: 
\begin{equation} \label{eqn:objective} \min_\Theta \mathbb{E}_{(X_i)\sim\text{Dataset}}\bigg[\mathcal{L}_{\text{MSE}}\bigg(f\big((X_i)_{i=1}^{n}  ; \Theta \big), (\mathbf{p}_{j \leftarrow i})_{i=1}^{n-1} \bigg)\bigg]
\end{equation}
At inference time, we let $(X_i)$ be the full US scan and reconstruct the tracking sequence $(T_i)$ as $T_i = T_0 \prod_{k=0}^{i-1}T_{k+1 \leftarrow k}$, with $T_{k+1\leftarrow k}$ converted from $\mathbf{p}_{k+1 \leftarrow k}$ estimated by the network. 
\noindent \textbf{Local Features Modeling:} Our local encoder is designed to capture local spatiotemporal features (e.g., speckle) from the ultrasound sequence. Following the literature, (e.g.~\cite{prevost20183d,guo2020sensorless,ning2022spatial,li2023trackerless}), it consists of a local spatiotemporal convolutional network followed by a spatial attention-pooling mechanism. Let $N, H$, and $W$ denote the number of timesteps, height and width of an ultrasound subsequence. Let $C$ and $C'$ denote hidden feature dimensions. First, a 3D CNN $g_\text{loc}: \mathbb{R}^{N \times H \times W}\rightarrow \mathbb{R}^{N \times \frac{H}{16} \times \frac{W}{16} \times C}$ extracts a sequence of feature maps from the sequence. Next, the spatial pooling module $h_\text{loc}: \mathbb{R}^{N \times \frac{H}{16} \times \frac{W}{16} \times C} \rightarrow \mathbb{R}^{N \times C'}$ extracts an embedding vector for each feature map in the sequence. The CNN is a 3D (two spatial and one time dimension) ResNet18 modified to have small temporal windows, ensuring the network captures only local temporal dependencies. The pooling module is a lightweight transformer~\cite{vaswani2017attention}. We train the network by attaching a linear layer and optimizing the tracking estimation objective (Eq.~\ref{eqn:objective}). We explicitly force locality by (i) using short contiguous subsquences as input; and (ii) keeping the image data at full resolution to preserve speckle features. 



\noindent \textbf{Global Features Modeling:}
Our global encoder is designed to extract coarse anatomical features from the ultrasound sequence. Its first stage is an image encoder $g_\text{glob}: \mathbb{R}^{N \times H \times W} \rightarrow \mathbb{R}^{N \times C'}$ which extracts a single C'-dimensional feature vector separately for each image in the input sequence, focusing on high-level features. We chose a simple 2D ResNet18 which worked well on this dataset but also explored transfer using pretrained backbones including the USFM~\cite{jiao2024usfm} encoder, the MedSAM~\cite{ma2024segment} encoder, and an iBOT~\cite{zhou2021ibot}-pretrained vision transformer. 
The second stage is a temporal self-attention module which aggregates information across time,  $h_\text{glob}: \mathbb{R}^{N \times C'} \rightarrow \mathbb{R}^{N \times C'}$. This module is based on the transformer~\cite{vaswani2017attention} architecture, whose powerful ability to capture long-term and global dependencies in sequences makes it a suitable choice for the global encoder. To learn the global features relevant to tracking estimation, we attach a linear layer to the global encoder and train it using the tracking estimation objective (Eq.~\ref{eqn:objective}). We explicitly encourage learning global context features by (i) using randomly sampled non-contiguous subsequences 
as inputs to the model, preventing it from leveraging short-term speckle motion and forcing it instead to rely on global positional awareness; (ii) feeding it with images downsampled to $224\times224$ to erase local textural details and allowing the model to focus on higher level semantics.


\noindent \textbf{Local-Global Fusion:} 
The fusion module combines embeddings from the local and global encoders. Following common multimodal late fusion approaches (e.g., ~\cite{lu2019vilbert,li2023blip}), we implement this fusion with a transformer decoder-based architecture using self-attention and cross-attention mechanisms. Specifically, local embeddings serve as the input states, while global embeddings act as the key-value states in the cross-attention layers. This design allows the transformer to refine local representations by attending to relevant global context, ensuring that short-term motion cues are interpreted within the broader anatomical and spatial structure of the scan. The fusion module is trained using the tracking estimation objective of Eq.~\ref{eqn:objective} with the full US sequence as input to the local encoder and an evenly spaced subsample\footnote{Subsampling improves inference speed and memory cost when using a large global encoder, at no observed performance cost. We typically subsample every 8 frames.} to the global encoder.

\section{Experiments}

\begin{figure}[t]
    \centering
    \includegraphics[width=1.0\linewidth]{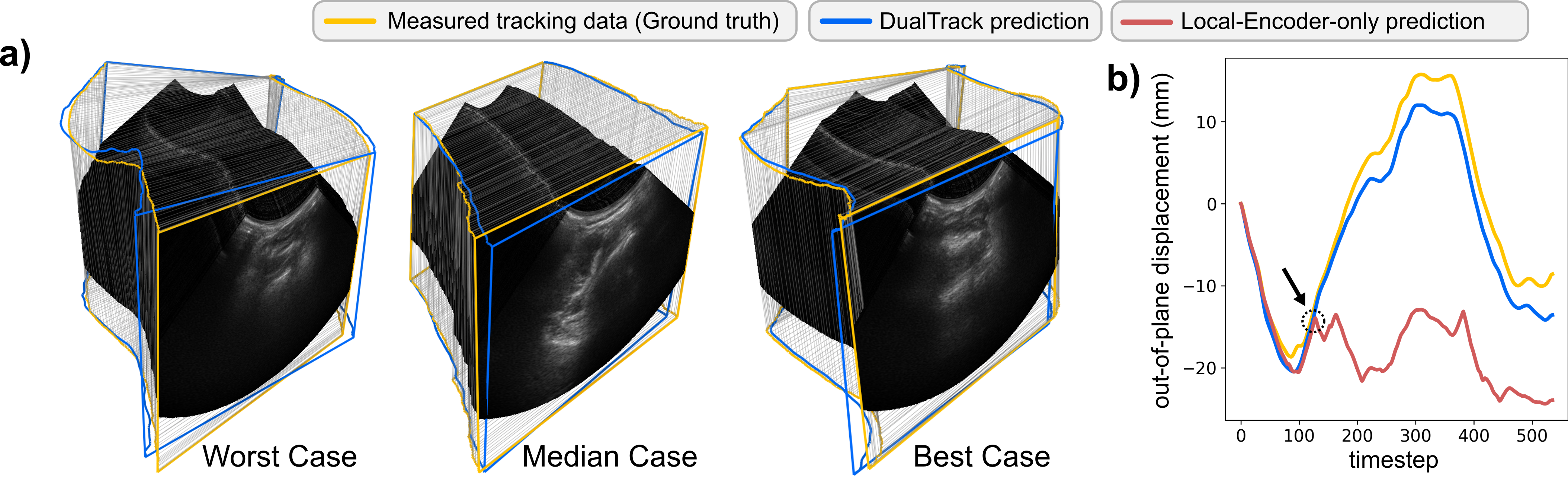}
    \caption{(a) Trajectories and 3D US images generated with \methodname{} predictions vs. ground truth, for three sweeps in the test set with worst/median/best GPE, respectively. (b) Comparison of the out-of-plane displacement prediction for the ``local only'' module and the full \methodname{} model. 
    The local model fails to disambiguate the out-of-plane direction and loses track after the first turn (see indicator); \methodname{} corrects this error using contextual cues from global features.}
    \label{fig:reconviz}
\end{figure}

To evaluate the \methodname{} approach, we performed the following experiments: 
\begin{enumerate}
    \item \emph{Comparison to prior art:} To determine the efficacy of our approach, we compare it to 2-Frame CNN~\cite{prevost20183d}, MoNet~\cite{luo2022deep}, DCL-net~\cite{guo2023} and Hybrid Transformer~\cite{ning2022spatial}. These works were chosen on the basis of (i) breadth of coverage of different methodologies in the literature; (ii) suitability of the method for our problem setting (i.e, no external sensor requirement); and (iii) success of reproduction. The methods were reproduced with reference to their manuscripts and, where available, source code. Modifications to the networks were made to account for differences in the dataset and hyperparameter values were re-tuned to allow fair comparison with our approach. 
    \item \emph{Ablation Studies:} To assess the contribution of each component, we conduct ablation studies by removing key components of our architecture. We test the performance of ``local only'', where we simply used our local encoder as the predictor, ``coupled local-global modeling'' where we add a temporal attention transformer on top of our local encoder (no decoupling), in comparison to our dual encoder decoupled approach. Additionally, we evaluate multiple image encoder backbones for the global context encoder.
\item \emph{Qualitative Analysis:} To further assess \methodname{}'s performance, we use it to reconstruct and visualize 
trajectories and 3D US based on the model's outputs (Figure~\ref{fig:reconviz}A). We specifically investigate the types of errors which occur in local predictions but are corrected by the global model.

    
    
\end{enumerate}

\noindent \textbf{Dataset:}
We use the public TUS-REC challenge dataset for this study~\cite{li2024trackerless,li2025tus}. It consists of 1,248 freehand 2D US scans of forearms from 53 volunteers, with multiple scan shapes, scan directions and probe orientations. Scan shapes included linear scans as well as scans with wavy, non-linear trajectories. The average translational distance from start to end of a scan is 167mm; the median scan length is 546 time steps (27 seconds). The total dataset size is approximately 226 GB. 
The public dataset is divided into training (first 50 subjects, 1176 scans) and test (last 3 subjects) splits. We further subdivided the training data into train (40 subjects) and validation (10 subjects) splits. 

\noindent \textbf{Evaluation strategy:} We evaluate our models by measuring reconstruction errors from predicted trajectories compared to ground truth: final drift rate (FDR) is the translational distance between the estimated and true position at the final timestep, normalized by the translational distance from start to end of a scan. Maximum drift is the largest absolute translational distance between the predicted and true position at any timestep during a scan. Following~\cite{li2024nonrigid}, we also use the \emph{point displacement errors} obtained by averaging the difference between the spatial positions of the five points (center and four corners) of an image when applying the true vs. predicted transform. The local point error (LPE) measures the average point error between two adjacent frames in the scan, while the global point error (GPE) measures the average point error between the true position and the fully reconstructed trajectory estimate at each timestep.
\noindent \textbf{Implementation Details:} Hyperparameters of our method were hand-tuned to minimize validation error. \textit{Training:} Models were implemented using PyTorch and trained using a Nvidia A40 GPU. Local encoder pretraining took approximately 7 days, divided into (i) 4000 epochs of training with the CNN encoder (learning rate (l.r.) $1e-4$, weight decay (w.d.) $1e-3$); and (ii) 500 epochs of training with the spatial attention and frozen CNN module (l.r. $1e-4$, w.d. $0$). Each training epoch for the local encoder sampled one subsequence of length 16 from each scan.
Global encoder pretraining took 24 hours (800 epochs), and fusion module training 48 hours for 500 epochs (both with l.r. $1e-4$, w.d. $0$). Baselines training took an average of 7 days to converge. All training phases used the AdamW optimizer and cosine annealed learning rate schedule. \textit{Architecture:} Transformer modules in the global encoder and fusion modules used a hidden size of 512 and intermediate size of 1024, 8 layers, and 8 attention heads. We found the inclusion of an additional small transformer (hidden size 64, intermediate size 32, 4 layers and attention heads) between the local encoder and fusion modules improved performance and training stability. 
 \textit{Inference:} On a single RTX Quatro 6000 GPU, DualTrack reconstructs a full sweep of 546 frames in 0.73 s, versus 0.52 s for Hybrid-Transformer and 0.46 s for MoNet, with a comparable memory footprint (< 7 GB). Given the median acquisition duration of 27 seconds, the extra 0.21 s inference time of our method is negligible. Most of the memory and time cost of inference is due to the local encoder CNN module, which could be replaced with a lighter-weight CNN if time compute resources were constrained.

\section{Results and Discussion}

\begin{table}[t]
    \centering
    \begin{tabular}{l|c|c|c|c|c|c}
    \hline
        \multicolumn{3}{c|}{\textbf{Method}} & \textbf{GPE (mm)} & \textbf{LPE ($\boldsymbol{\mu}$m)} & \textbf{FDR (\%)} & \textbf{Max. Drift (mm)} \\
        \hline
         \multicolumn{7}{c}{} \\
        \multicolumn{7}{c}{\textbf{Baseline Comparison}} \\
        \hline
        \multicolumn{3}{c|}{DCL-net~\cite{guo2020sensorless}} & 10.77 & 134.21 & 11.68 & 18.75 \\
        \multicolumn{3}{c|}{2-Frame CNN~\cite{prevost20183d}} & 8.90 & 140.21 & 8.93 & 14.49 \\ 
        \hline
        \multicolumn{3}{c|}{MoNet~\cite{luo2022deep}} & 9.19 & 127.97 & 8.87 & 14.46\\ 
        \multicolumn{3}{c|}{Hybrid Transformer~\cite{ning2022spatial}} & 6.01 & 122.49 & 6.22 & 10.00 \\ 
        \hline
        \multicolumn{3}{c|}{\methodname{} (ours)} & \textbf{4.93} & \textbf{122.01} & \textbf{5.10} & \textbf{8.33} \\
        \hline 
        \multicolumn{7}{c}{} \\
        \multicolumn{7}{c}{\textbf{Ablation Results}} \\
        Local & Global & Dual Encoder \\
        \hline
        \xmark & \cmark & \xmark &  11.26 & 757.64 &   12.48 & 19.73 \\
         \cmark  & \xmark & \xmark & 7.36 & 140.90 & 6.96 & 11.70\\ 
        \cmark & \cmark & \xmark & 5.92 & 135.32 &  6.17 & 9.77\\
        \hline
       \cmark & \cmark & \cmark & \textbf{4.93} & \textbf{122.01} & \textbf{5.10} & \textbf{8.33} \\
        \hline
        \multicolumn{7}{c}{} \\
    \end{tabular}
    \caption{Averaged reconstruction metrics on 72 public test scans. For all metrics, lower is better with best results in bold. \methodname{} (our method) outperforms the literature across all metrics, while ablation studies highlight the importance of each component.}
    \label{tab:comparison}
\end{table}

\begin{figure}[t]
    \centering
    \includegraphics[width=1.0\linewidth]{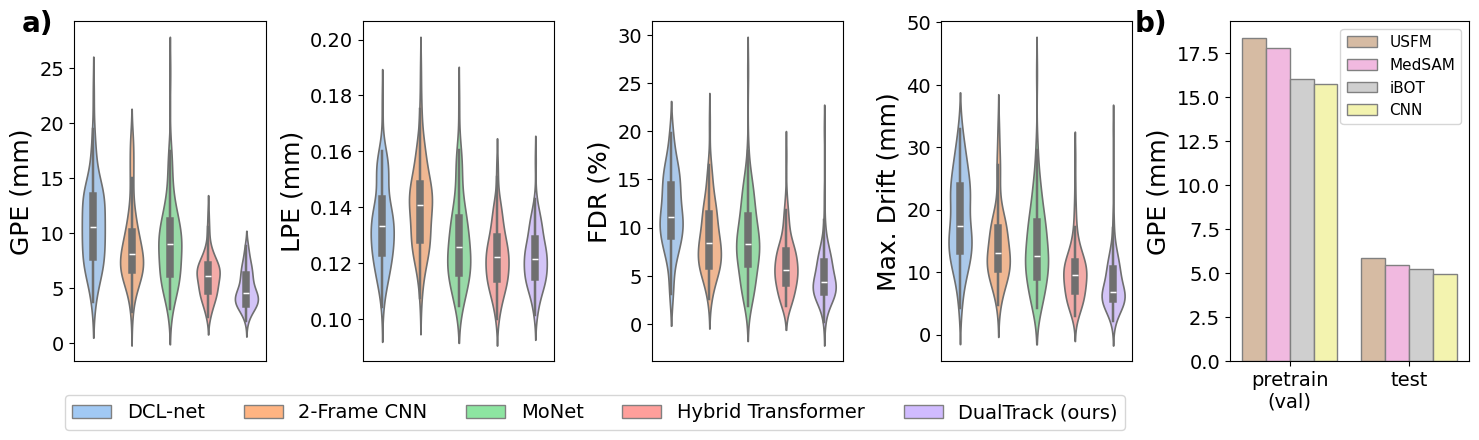}
    \caption{(a) Distribution of tracking reconstruction errors across the 72 test scans for \methodname{} and competing methods. Our method significantly ($p \leq 0.005$) outperforms its leading competitor on all metrics. (b) DualTrack can be successfully used with a variety of image backbones, giving the flexibility to leverage transfer learning strategies.
    }
    \label{fig:boxplots}
\end{figure}

\methodname{} outperforms prior methods across all metrics. Table~\ref{tab:comparison} (top) reports averaged error metrics on the TUS-REC test set. Compared to Hybrid Transformer, its leading competitor, \methodname{} achieves lower errors (-1.08~$mm$ GPE, -0.48~$\mu m$ LPE, -1.12\% FDR, -1.67~$mm$ Max. Drift). Figure~\ref{fig:boxplots}a shows error distributions across scans, with \methodname{} exhibiting consistently lower values for key global metrics (GPE, LPE, FDR). These improvements were statistically significant (Wilcoxon signed-rank, $p \leq 0.005$). Models relying solely on local features (Rows 1-2) were outperformed by those using global features (Rows 3-5). The additional performance gains of \methodname{} over other global-aware models (Rows 3-4) likely stem from its dual encoder architecture, which explicitly separates local and global modeling for more robust feature extraction and integration.

Results of our ablation study (Table~\ref{tab:comparison}, bottom) show the importance of both global feature extraction (improvements from Row 2 to 3), and the dual encoder setup (improvements from Row 3 to 4), further highlighting the strength of \methodname{}'s design. Figure~\ref{fig:boxplots}b shows the impact of different global encoder backbones on GPE. In general, all backbones performed well, indicating the flexibility of the dual encoder. A simple 2D CNN gave the best performance on this dataset, followed by the iBOT self-supervised model. We hypothesize that the best choice of backbone is likely to be dataset-specific with (i) smaller datasets, (ii) datasets featuring more recognizable anatomical structures than the forearm, or (iii) datasets with organs that were represented in a foundation model's training corpus.
Further studies are needed to test this hypothesis.

Visualization of trajectories and 3D US reconstructions generated by \methodname{} show notable quality. We highlight visualizations on three test cases (Figure~\ref{fig:reconviz}a) which had the worst, median, and best GPE respectively. The model's prediction aligns with ground truth. Even the worst case followed the general trajectory closely with a slight angular offset at the end of the scan, with a GPE of only 8~$mm$--the best case had a GPE less than 2~$mm$. A fourth case (Figure~\ref{fig:reconviz}B) was chosen to highlight the positive impact of global modeling by comparing the out-of-plane displacement prediction for the ``local only'' model vs. the full \methodname{} model. Unlike \methodname{}, the local model fails to disambiguate the out-of-plane direction and loses track after the first turn (see indicator).

\section{Conclusion} 
In this work, we addressed the need for robust local and global feature modeling in sensorless 3D US reconstruction. We introduced \methodname{}, a dual encoder architecture that explicitly separates fine-grained features from high-level anatomical cues.
Experimental results demonstrate that this decoupled design significantly reduces reconstruction errors compared to existing state-of-the-art. 
Given its low error rate, \methodname{} is likely capable of generating 3D ultrasound volumes that meet the accuracy requirements of clinical applications, especially those involving short and steady scans such as carotid plaque quantification: further investigation into these applications should be the subject of future work. Furthermore, future work should investigate the efficacy of DualTrack for complex freehand scans of anatomically variable regions such as the liver or heart.

\begin{credits}
\subsubsection{\ackname} This study was partially funded by the Accelerate program of Mitacs Inc., Canada.
\end{credits}

%
%
%
\bibliographystyle{splncs04}
\bibliography{bibliography}

@article{prevost20183d,
  title={3D freehand ultrasound without external tracking using deep learning},
  author={Prevost, Raphael and Salehi, Mehrdad and Jagoda, Simon and Kumar, Navneet and Sprung, Julian and Ladikos, Alexander and Bauer, Robert and Zettinig, Oliver and Wein, Wolfgang},
  journal={Medical image analysis},
  volume={48},
  pages={187--202},
  year={2018},
  publisher={Elsevier}
}

@inproceedings{guo2020sensorless,
  title={Sensorless freehand 3D ultrasound reconstruction via deep contextual learning},
  author={Guo, Hengtao and Xu, Sheng and Wood, Bradford and Yan, Pingkun},
  booktitle={Medical Image Computing and Computer Assisted Intervention--MICCAI 2020: 23rd International Conference, Lima, Peru, October 4--8, 2020, Proceedings, Part III 23},
  pages={463--472},
  year={2020},
  organization={Springer}
}

@ARTICLE{guo2023,
  author={Guo, Hengtao and Chao, Hanqing and Xu, Sheng and Wood, Bradford J. and Wang, Jing and Yan, Pingkun},
  journal={IEEE Transactions on Biomedical Engineering}, 
  title={Ultrasound Volume Reconstruction From Freehand Scans Without Tracking}, 
  year={2023},
  volume={70},
  number={3},
  pages={970-979},
  doi={10.1109/TBME.2022.3206596}
}

@inproceedings{luo2022deep,
  title={Deep motion network for freehand 3D ultrasound reconstruction},
  author={Luo, Mingyuan and Yang, Xin and Wang, Hongzhang and Du, Liwei and Ni, Dong},
  booktitle={International Conference on Medical Image Computing and Computer-Assisted Intervention},
  pages={290--299},
  year={2022},
  organization={Springer}
}

@inproceedings{ning2022spatial,
  title={Spatial position estimation method for 3d ultrasound reconstruction based on hybrid transfomers},
  author={Ning, Guochen and Liang, Hanying and Zhou, Lei and Zhang, Xinran and Liao, Hongen},
  booktitle={2022 IEEE 19th International Symposium on Biomedical Imaging (ISBI)},
  pages={1--5},
  year={2022},
  organization={IEEE}
}

@inproceedings{li2023trackerless,
  title={Trackerless freehand ultrasound with sequence modelling and auxiliary transformation over past and future frames},
  author={Li, Qi and Shen, Ziyi and Li, Qian and Barratt, Dean C and Dowrick, Thomas and Clarkson, Matthew J and Vercauteren, Tom and Hu, Yipeng},
  booktitle={2023 IEEE 20th International Symposium on Biomedical Imaging (ISBI)},
  pages={1--5},
  year={2023},
  organization={IEEE}
}

@inproceedings{luo2021,
author = {Luo, Mingyuan and Yang, Xin and Huang, Xiaoqiong and Huang, Yuhao and Zou, Yuxin and Hu, Xindi and Ravikumar, Nishant and Frangi, Alejandro F. and Ni, Dong},
title = {Self Context and Shape Prior for Sensorless Freehand 3D Ultrasound Reconstruction},
year = {2021},
isbn = {978-3-030-87230-4},
url = {https://doi.org/10.1007/978-3-030-87231-1_20},
doi = {10.1007/978-3-030-87231-1_20},
booktitle = {Medical Image Computing and Computer Assisted Intervention – MICCAI 2021: 24th International Conference, Strasbourg, France, September 27–October 1, 2021, Proceedings, Part VI}
}

@article{vaswani2017attention,
  title={Attention is all you need},
  author={Vaswani, Ashish and Shazeer, Noam and Parmar, Niki and Uszkoreit, Jakob and Jones, Llion and Gomez, Aidan N and Kaiser, {\L}ukasz and Polosukhin, Illia},
  journal={Advances in neural information processing systems},
  volume={30},
  year={2017}
}

@article{zhou2021ibot,
  title={ibot: Image bert pre-training with online tokenizer},
  author={Zhou, Jinghao and Wei, Chen and Wang, Huiyu and Shen, Wei and Xie, Cihang and Yuille, Alan and Kong, Tao},
  journal={arXiv preprint arXiv:2111.07832},
  year={2021}
}

@inproceedings{li2024trackerless,
  author       = {Li, Q. and Saeed, S. U. and Barratt, D. C. and Clarkson, M. J. and Vercauteren, T. and Hu, Y.},
  title        = {Trackerless 3D Freehand Ultrasound Reconstruction Challenge},
  booktitle    = {27th International Conference on Medical Image Computing and Computer Assisted Intervention (MICCAI 2024)},
  year         = {2024},
  publisher    = {Zenodo},
  doi          = {10.5281/zenodo.10991501},
  url          = {https://doi.org/10.5281/zenodo.10991501}
}

@article{ma2024segment,
  title={Segment anything in medical images},
  author={Ma, Jun and He, Yuting and Li, Feifei and Han, Lin and You, Chenyu and Wang, Bo},
  journal={Nature Communications},
  volume={15},
  number={1},
  pages={654},
  year={2024},
  publisher={Nature Publishing Group UK London}
}

@article{jiao2024usfm,
  title={Usfm: A universal ultrasound foundation model generalized to tasks and organs towards label efficient image analysis},
  author={Jiao, Jing and Zhou, Jin and Li, Xiaokang and Xia, Menghua and Huang, Yi and Huang, Lihong and Wang, Na and Zhang, Xiaofan and Zhou, Shichong and Wang, Yuanyuan and others},
  journal={Medical Image Analysis},
  volume={96},
  pages={103202},
  year={2024},
  publisher={Elsevier}
}

@inproceedings{li2024nonrigid,
  title={Nonrigid Reconstruction of Freehand Ultrasound Without a Tracker},
  author={Li, Qi and Shen, Ziyi and Yang, Qianye and Barratt, Dean C and Clarkson, Matthew J and Vercauteren, Tom and Hu, Yipeng},
  booktitle={International Conference on Medical Image Computing and Computer-Assisted Intervention},
  pages={689--699},
  year={2024},
  organization={Springer}
}

@article{hong2023dual,
  title={Dual encoder network with transformer-CNN for multi-organ segmentation},
  author={Hong, Zhifang and Chen, Mingzhi and Hu, Weijie and Yan, Shiyu and Qu, Aiping and Chen, Lingna and Chen, Junxi},
  journal={Medical \& biological engineering \& computing},
  volume={61},
  number={3},
  pages={661--671},
  year={2023},
  publisher={Springer}
}

@inproceedings{zhang2021transfuse,
  title={Transfuse: Fusing transformers and cnns for medical image segmentation},
  author={Zhang, Yundong and Liu, Huiye and Hu, Qiang},
  booktitle={Medical image computing and computer assisted intervention--MICCAI 2021: 24th international conference, Strasbourg, France, September 27--October 1, 2021, proceedings, Part I 24},
  pages={14--24},
  year={2021},
  organization={Springer}
}

@article{fu2022deau,
  title={DEAU-Net: Attention networks based on dual encoder for Medical Image Segmentation},
  author={Fu, Zhaojin and Li, Jinjiang and Hua, Zhen},
  journal={Computers in Biology and Medicine},
  volume={150},
  pages={106197},
  year={2022},
  publisher={Elsevier}
}

@inproceedings{farshad2022net,
  title={Y-net: A spatiospectral dual-encoder network for medical image segmentation},
  author={Farshad, Azade and Yeganeh, Yousef and Gehlbach, Peter and Navab, Nassir},
  booktitle={International conference on medical image computing and computer-assisted intervention},
  pages={582--592},
  year={2022},
  organization={Springer}
}

@article{lu2019vilbert,
  title={Vilbert: Pretraining task-agnostic visiolinguistic representations for vision-and-language tasks},
  author={Lu, Jiasen and Batra, Dhruv and Parikh, Devi and Lee, Stefan},
  journal={Advances in neural information processing systems},
  volume={32},
  year={2019}
}

@inproceedings{li2023blip,
  title={Blip-2: Bootstrapping language-image pre-training with frozen image encoders and large language models},
  author={Li, Junnan and Li, Dongxu and Savarese, Silvio and Hoi, Steven},
  booktitle={International conference on machine learning},
  pages={19730--19742},
  year={2023},
  organization={PMLR}
}

@article{li2025tus,
  title={TUS-REC2024: A challenge to reconstruct 3D freehand ultrasound without external tracker},
  author={Li, Qi and Saeed, Shaheer U and Huang, Yuliang and Luo, Mingyuan and Yan, Zhongnuo and Chen, Jiongquan and Yang, Xin and Ni, Dong and Winter, Nektarios and Nguyen, Phuc and others},
  journal={arXiv preprint arXiv:2506.21765},
  year={2025}
}
%




\end{document}